\documentclass[journal]{IEEEtran}
\usepackage{makecell}
\usepackage{multirow}

\usepackage{epsfig}
\usepackage{graphicx}
\usepackage{amsmath,amssymb} % define this before the line numbering.
\usepackage{color}
\usepackage[switch,mathlines]{lineno}
\usepackage{array}
\usepackage{bbm}
\usepackage{subcaption}
\usepackage{colortbl}
\usepackage[pagebackref=true,breaklinks=true,colorlinks,bookmarks=false]{hyperref}
\usepackage{verbatim}
\makeatletter
\DeclareRobustCommand\onedot{\futurelet\@let@token\@onedot}
\def\@onedot{\ifx\@let@token.\else.\null\fi\xspace}

\makeatother

\begin{document}
%\linenumbers

\title{NPCFace: Negative-Positive Collaborative Training for Large-scale Face Recognition}

\author{Dan Zeng,~\IEEEmembership{Member, IEEE},
        Hailin Shi,~\IEEEmembership{Member, IEEE},
        Hang Du,~\IEEEmembership{}
        Jun Wang,~\IEEEmembership{}
        \\
        Zhen Lei,~\IEEEmembership{Senior Member, IEEE}
        and Tao Mei~\IEEEmembership{Fellow, IEEE} 
        % <-this % stops a space
\thanks{D. Zeng and H. Du are with the Key Laboratory of Specialty Fiber Optics and Optical Access Networks, Joint International Research Laboratory of Specialty Fiber Optics and Advanced Communication, Shanghai Institute of Advanced Communication and Data Science, Shanghai University, Shanghai, 200444, China (e-mail: dzeng@shu.edu.cn, duhang@shu.edu.cn)}
\thanks{H. Shi, J. Wang and T. Mei are with the JD AI Research, Beijing, 100029, China (e-mail: shihailin@jd.com, wangjun492@jd.com, tmei@live.com)}
\thanks{Z. Lei is with the National Laboratory of Pattern Recognition, Institute of Automation, Chinese Academy of Science, Beijing, 100190, China (e-mail: zlei@nlpr.ia.ac.cn)}

}

%{Shell \MakeLowercase{\etal}:}

\maketitle

\begin{abstract}
The training scheme of deep face recognition has greatly evolved in the past years, yet it encounters new challenges in the large-scale data situation where massive and diverse hard cases occur.
% problem
Especially in the range of low false accept rate (FAR), there are various hard cases in both positives (intra-class) and negatives (inter-class). 
% our objective
In this paper, we study how to make better use of these hard samples for improving the training.
% previous solution
The literature approaches this by margin-based formulation in either positive logit %(such as SphereFace, CosFace, ArcFace) 
or negative logits. %(such as MV-softmax, ArcNegFace, CurricularFace).
% their flaws
However, the correlation between hard positive and hard negative is overlooked, and so is the relation between the margins in positive and negative logits.
We find such correlation is significant, especially in the large-scale dataset, and one can take advantage from it to boost the training via relating the positive and negative margins for each training sample.
% our goal
To this end, we propose an explicit collaboration between positive and negative margins sample-wisely.
% our method
Given a batch of hard samples, a novel Negative-Positive Collaboration loss, named NPCFace, is formulated, which emphasizes the training on both negative and positive hard cases via the collaborative-margin mechanism in the softmax logits, and also brings better interpretation of negative-positive hardness correlation.
Besides, the emphasis is implemented with an improved formulation to achieve stable convergence and flexible parameter setting. 
% problem to cope with
% conclusion
We validate the effectiveness of our approach on various benchmarks of large-scale face recognition, and obtain advantageous results especially in the low FAR range.
\end{abstract}


% \fi
\IEEEpeerreviewmaketitle

\section{Introduction}
\label{sec_intro}
%\IEEEPARstart{F}{ace} recognition is xxxx.

Face recognition is a long-standing topic in computer vision and video analysis. With the advances of deep learning for face recognition~\cite{taigman2014deepface,sun2014deep,schroff2015facenet}, increasing research interest focuses on the large-scale face recognition whose major challenge falls in the recognition accuracy at the low false accept rate (FAR) range. 
There are various factors leading to diverse hard positive (intra-class)  cases at the low FAR, such as large pose, age gap, non-uniform lightening, occlusion, and so forth; besides, the similar-looking factor, in which two identities have similar faces, leads to many hard negative (inter-class) cases.
In large-scale data, the hard positive and hard negative co-occur much frequently, which make face recognition more difficult.

%These hard cases form not only hard positives but also hard negatives. 
%Here, positive and negative denote intra-class and inter-class, respectively.
Many prior methods~\cite{schroff2015facenet,duan2019deep,zhou2017efficient,wu2017sampling} aims to select training samples from the hard cases to gain performance improvement.
Rather than study how to identify hard samples from the dataset, in this paper, given a definition of hard sample (such as mis-classified sample), the goal is to study how to make better use of them to boost the training.

\par
 
Recently, many methods are proposed to establish the training supervision from the perspective of either positive or negative, and achieve great progress on the mainstream benchmarks.
Some of them~\cite{liu2016large,liu2017sphereface,wang2018cosface,wang2018additive,deng2019arcface,liu2019adaptiveface} aim to enlarge the gap between different classes by adding an angular margin in the positive logit of softmax. Liu \textit{et al.}~\cite{liu2016large,liu2017sphereface} introduces the idea of angular margin at the first time. CosFace~\cite{wang2018cosface} and AM-softmax~\cite{wang2018additive} propose an additive margin for better optimization. ArcFace~\cite{deng2019arcface} improves the geometric interpretation of the margin and achieves better performance. AdaptiveFace~\cite{liu2019adaptiveface} encourages to learn an adaptive margin for each class. The above methods can be regarded as the category of hard positive mining, because they aim to emphasize the training on those samples away from their ground-truth center by adding margin in the positive logit of softmax. In contrast, some other methods~\cite{wang2019mis,liu2019towards,Huang_2020_CVPR} consider to employ hard negative mining by adding margin from the negative (non-ground-truth) view. MV-softmax~\cite{wang2019mis} proposes to identify the mis-classified samples and exploit them by adding margin in the negative logits. 
%It can be integrated to other positive-view methfods (\ie AM-softmax~\cite{wang2018additive} and ArcFace~\cite{deng2019arcface}), while there is no collaboration between positive and negative. 
ArcNegFace~\cite{liu2019towards} also studies on margin-involved negative logits in a similar way. Based on MV-softmax, CurricularFace~\cite{Huang_2020_CVPR} adaptively adjusts the relative importance of easy and hard samples during different training stage by modulating the negative logits. A brief summary of them can be found in Table~\ref{existing_methods}.

The above-mentioned methods study to improve the training from either the positive view or negative view. To further make full use of the hard samples from positive perspective and negative perspective simultaneously, a straightforward idea is to add margins in both positive and negative logits, such as the manner in ~\cite{wang2019mis,liu2019towards,Huang_2020_CVPR}.
However, such straight combination has a shortcoming: the margins are imposed independently in positive and negative logits, which is a sub-optimal choice of setting the margin for each hard sample in training. We argue that the margins should be related between negative and positive in a collaborative way, sample-wisely. 
The reason is that, given a face dataset, a sample, which acts as a hard positive, will generally act as a hard negative as well. Such case widely exists in the face dataset, especially when the dataset is of large-scale. For example, as shown in Fig.~\ref{hard_sample_scheme} (b), when dataset is of large-scale, 
``Class 1'' is surrounded by many neighboring classes, and the hard positive sample could easily find a neighboring class to form a hard negative. More formally, the hard positives and hard negatives have significant correlation in large-scale face dataset. This phenomenon is verified in the datasets CASIA-WebFace~\cite{yi2014learning} and MS-Celeb-1M~\cite{guo2016ms} in Section~\ref{neg and pos correlation} (Fig.~\ref{coorealtion}). 
This observation is consistent with the intuition, but has been overlooked in the prior methods for face recognition.

\begin{figure}[t]
\centering
\includegraphics[scale=0.3]{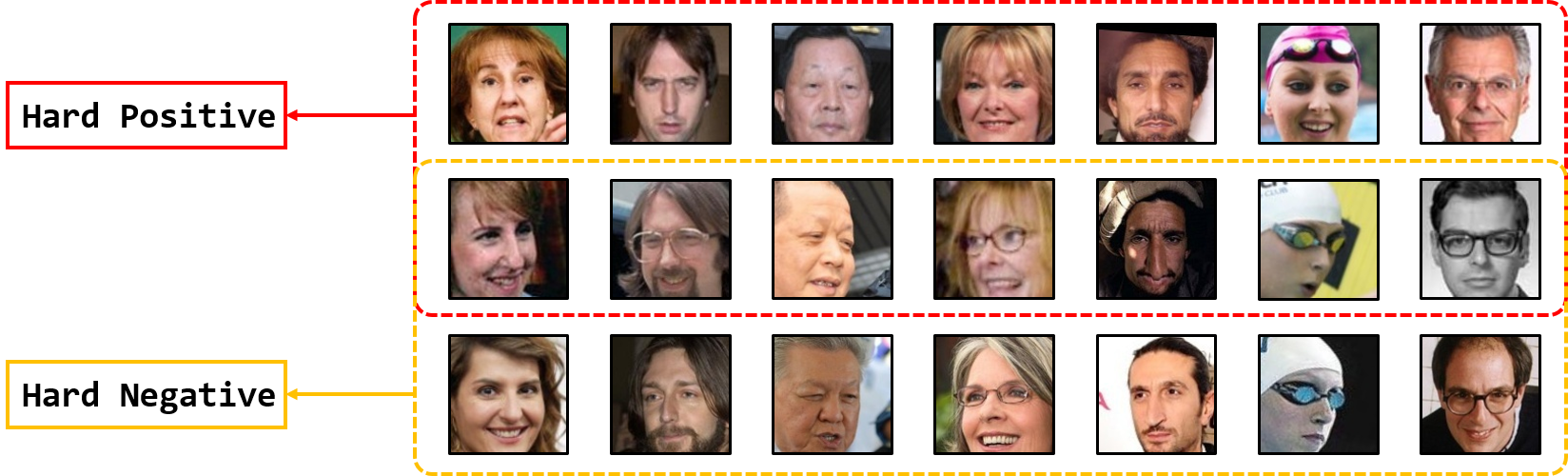}
\caption{The top and middle form hard positives (a.k.a. intra-class), the middle and bottom form hard negatives (a.k.a. inter-class). 
The co-occurrence of hard positive and hard negative is frequent in large-scale dataset, since similar looking person can be easily found in large population.}
\label{hard_pairs}
\end{figure}

%------------------------------------------------------------------------
To address this issue, we propose the Negative-Positive Collaboration loss (NPCFace): 
applying the hard negative-positive correlation to the training loss formulation, 
and so taking the benefit of it via a collaborative margin scheme for better supervision. 
Specifically, we formulate an explicit collaboration of the margins in the positive and negative logits. The margin in the positive logit will be enlarged by the collaboration if the negative logit is enlarged. This collaboration scheme is implemented sample-wisely: it will be activated when the sample is identified as a hard sample by an off-the-shelf criterion; otherwise, the collaboration will be deactivated, and the positive logit and negative logit will be calculated independently. 
Through this collaboration scheme, NPCFace emphasizes the supervision of the hard samples from both the positive and negative perspective.
A training sample, which acts as a hard case from the negative view, will also give extra contribution to the supervision from the positive view.
Resorting to the collaboration scheme, our NPCFace achieves better exploitation of large-scale training data of face recognition, and pushes the frontier in low FAR range.

Furthermore, we improve the margin formulation in the negative logits in order to guarantee the stable convergence and flexible parameter setting. The experiments on different network architectures and training datasets show that one can easily adopt our method to conduct effective training and further study.

The contributions of this paper are summarized as follows:
\begin{itemize}
\item We propose a novel supervision loss, named NPCFace, to improve the usage of hard samples in the large-scale training. It performs a collaborative training emphasis on hard positives and hard negatives sample-wisely, and is implemented via an explicitly related margin formulation in the softmax logits. 
Benefiting from the correlation between hard positive and hard negative, NPCFace makes better use of the hard samples for training.
\item We improve the margin formulation in the negative logits to achieve stable convergence and flexible parameter setting. These benefits are validated on various network architectures and face datasets. One can easily train the deep networks with NPCFace on large-scale face datasets.
\item We evaluate our approach on extensive face recognition benchmarks, including LFW, BLUFR, CALFW, CPLFW, CFP-FP, AgeDB-30, RFW, IJB-B, IJB-C, MegaFace and Trillion-Pairs. Resorting to the above improvements, NPCFace achieves leading performance on them, especially in the low FAR range.
\item NPCFace also shows advantage for generic object recognition. The results on image retrieval datasets can be found in the supplemental material. 
\end{itemize}

\begin{table}[t]
\begin{center}
\caption{Representative methods for margin-based formulation in positive and negative logits.}
\label{existing_methods}
%\vspace{0.2cm}
\begin{tabular}{|p{2.3cm}<{\centering}|p{5.4cm}<{\centering}|}
\hline
Margin formulation & Methods \\
\hline
Positive logit & SphereFace, CosFace, ArcFace, \textit{etc.} \\
\hline
Negative logits & MV-softmax, ArcNegFace, CurricularFace, \textit{etc.} \\
\hline
\end{tabular}
\end{center}
\end{table}

\section{Related Work}
\label{sec_related}

\subsection{Loss Function}
\label{subsec_related_loss}
Loss function is an essential research topic in deep supervision for face recognition. There are mainly two routines in the previous works. 
The first consists in explicit metric learning that focuses on optimizing the feature distance according to the sample-relative labels. 
Contrastive loss~\cite{chopra2005learning,hadsell2006dimensionality,sun2014deep} calculates pairwise Euclidean distance and optimizes it in feature space, while Triplet loss~\cite{schroff2015facenet} selects the triplet samples and measures the relative Euclidean distance of them.
The second includes classification loss functions, which allows to learn discriminative face feature embedding via classification learning.
For example, Taigman \textit{et al.}~\cite{taigman2014deepface} which aims to make the different identities separate. Furthermore, face feature representation should be compact in intra-class and separate in inter-class simultaneously. Therefore, Center loss~\cite{wen2016discriminative} develops a method to constrain the intra-class compactness. RegularFace~\cite{zhao2019regularface} aims at enlarging inter-class separability between different class centers by an exclusive regularization.  L-softmax~\cite{liu2016large} and SphereFace~\cite{liu2017sphereface} introduce the angular margin to obtain significant improvement.
%, but the multiplicative margin needs approximated calculation and careful adjustment. 
NormFace~\cite{wang2017normface} studies the effectiveness of the feature and weight normalization.
Afterward, CosFace~\cite{wang2018cosface} and AM-softmax~\cite{wang2018additive} propose an additive margin to the positive logit which can be optimized steadily. ArcFace~\cite{deng2019arcface} employs an additive angular margin, which has a more clear geometric interpretation and achieves further improvement. AdaCos~\cite{zhang2019adacos} introduces an adaptive scale parameter to reformulate the mapping between cosine similarity and classification probability.
% Wu \etal~\cite{Wu_2020_CVPR} presents a rotation consistent margin (RCM) loss to keep the intra-class compactness of low-bit face recognition models. 
More recently, MV-softmax~\cite{wang2019mis}, ArcNegFace~\cite{liu2019towards}, CurricularFace~\cite{Huang_2020_CVPR} propose to add margins in the negative logits. However, seldom has yet accomplished thorough study on the collaboration between positive and negative.

\subsection{Hard Sample Usage}
\label{subsec_related_mining}
There are many prior works that study the mining approach for hard samples, such as OHEM~\cite{shrivastava2016training}, SmartMining~\cite{harwood2017smart}, HDC~\cite{yuan2017hard} and some others~\cite{duan2019deep,zhou2017efficient,wu2017sampling} for face and general learning. However, there are fewer literatures of discussion about how to use the selected hard samples. 
FaceNet~\cite{schroff2015facenet} selects hard positive samples and semi-hard negative samples to construct the triplets.
EDM~\cite{shi2016embedding} proposes to use the moderate positive and hard negative in a related manner. 
MV-softmax~\cite{wang2019mis} chooses the mis-classified sample as hard one and enlarges the corresponding loss value by adjusting the margins explicitly. 
Similarly, ArcNegFace~\cite{liu2019towards} proposed a margin-involved negative logits to emphasis the hard samples. 
Based on MV-softmax, CurricularFace~\cite{Huang_2020_CVPR} adaptively adjusts the relative importance of easy and hard samples during different training stages by modulating the negative logits. 
Earlier, a series of methods propose to exploit the hard samples in a more implicit way. These methods, such as CosFace~\cite{wang2018cosface}, AM-softmax~\cite{wang2018additive}, ArcFace~\cite{deng2019arcface},  
mainly adopt the margins in loss function, so the training supervision automatically focuses on the hard samples.
AdaptiveFace~\cite{liu2019adaptiveface} makes the model to learn the adaptive margin for each class and focus on hard classes with a small number of hard prototypes in each training iteration.

\begin{figure}[t]
     %\centering
     \hspace{-10pt}
     \begin{subfigure}[t]{0.24\textwidth}
         \centering
         \includegraphics[height=4.2cm]{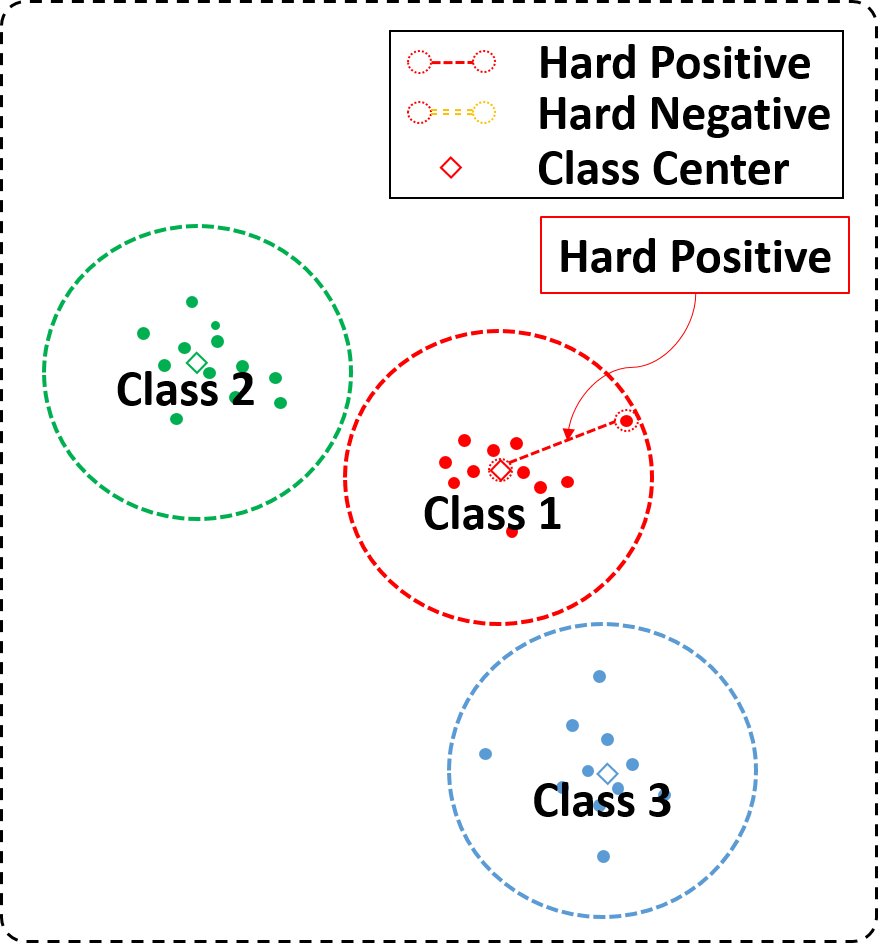}
         \caption{}
     \end{subfigure}
     %\hfill
     \hspace{-10pt}
     \begin{subfigure}[t]{0.24\textwidth}
         \centering
         \includegraphics[height=4.2cm]{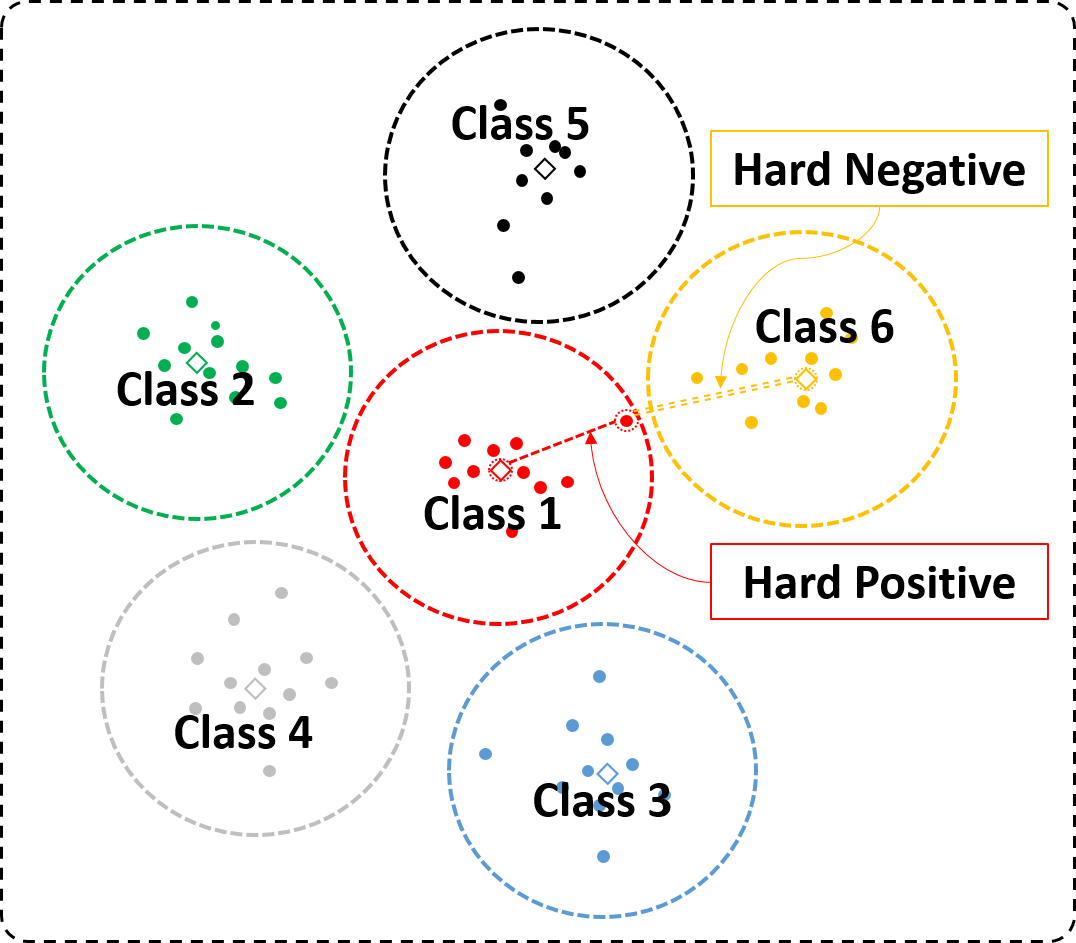}
         \caption{}
     \end{subfigure}

    \caption{
    (a) An example of hard positive in small-scale dataset.
    (b) In large-scale dataset, there is high possibility of co-occurrence of hard positive and hard negative. This is also verified by high correlation computed in Section~\ref{neg and pos correlation} (Fig.~\ref{coorealtion}). Best viewed in color.}
        \label{hard_sample_scheme}
\end{figure}

\section{Our Approach}
\label{sec_method}

\subsection{Revisiting Softmax}
\label{subsec_revisit}
The softmax loss is the most widely used training loss function, which includes a fully connected layer, the softmax function and the cross-entropy loss. At the fully connected layer, the output ${w}_{j}^{\mathrm{T}} {x}_{i}$ is obtained by the inner product of the ${i}$-th feature ${x}_{i}$ and the ${j}$-th class weight ${w}_{j}$. After $l_2$ normalization on features and weights, the inner product equals to the cosine similarity ${w}_{j}^{\mathrm{T}}{x}_{i}=\cos{\left(\theta_{{i},{j}}\right)}$. Thus, the softmax loss can be formulated as follows:
 
\begin{equation}
\mathcal{L} = -\frac{1}{N} \sum_{i=1}^{N}\log \frac{e^{s \cos \left(\theta_{{i},{y}}\right)}}{e^{s \cos \left(\theta_{{i},{y}}\right)}+\sum_{j=1,j\neq y}^{C} e^{s \cos \left(\theta_{{i},{j}}\right)}},
\label{original softmax}
\end{equation}
where ${N}$ is the batch size, ${C}$ is the class number, $s$ is a re-scaling parameter, and $y$ is the ground-truth label of the ${i}$-th sample. We denote the positive and negative logits as $f_y$ and $f_j$, which are computed as $f_y = s \cos \left(\theta_{{i},{y}}\right)$ and $f_j = s \cos \left(\theta_{{i},{j}}\right) | _{j \neq y}$, respectively. So, the softmax loss can be further formulated as: 
 
\begin{equation}
    \mathcal{L} = -\frac{1}{N} \sum_{i=1}^{N}\log \frac{e^{f_y}}
    {e^{f_y}+\sum_{j=1,j\neq y}^{C} e^{f_j}}.
\label{softmax_2}
\end{equation}
Then, the gradient with respect to the positive logit and negative logits is calculate as:
 
\begin{equation}
    \frac{\partial \mathcal{L}}{\partial f_k} = 
    \begin{cases}
     p_k - 1 & \text{if } k = y, \\
     p_k & \text{if } k \neq y,
    \end{cases}
\label{gradient}
\end{equation}
where $p_k$ is the predicted probability on the $k$-th class, which is defined by the softmax function:
 
\begin{equation}
    p_k = \frac{e^{f_k}}{\sum_{j=1}^{C} e^{f_j}}.
\label{softmax_function}
\end{equation}
Given $\sum_{k}^{C} p_k = 1$ for the total $C$ classes, the gradient summation of each sample with respect to each class always equals to the constant zero: 
 
\begin{equation}
    \sum_{k}^{C} \frac{\partial \mathcal{L}}{\partial f_k} = 0.
\label{sum_gradient}
\end{equation}
Considering $p_k$ is a probability that being less than 1, the gradient with respect to the positive logit ($p_k -1$) and that with respect to the negative logit ($p_k$) have the opposite sign. 
Therefore, for each training sample, given the loss function of softmax, the supervisions on the ground-truth class and non-ground-truth class have strong correlation in terms of magnitude, since their sum equals to zero.
In other words, if a training sample leads to a strong supervision on the ground-truth class, then it will bring strong supervision on the non-ground-truth classes as well. 
This property is brought by the nature of softmax function, which normalizes the sum of the output logits to $1$.

\begin{figure}[t]
\centering
\includegraphics[height=3.8cm]{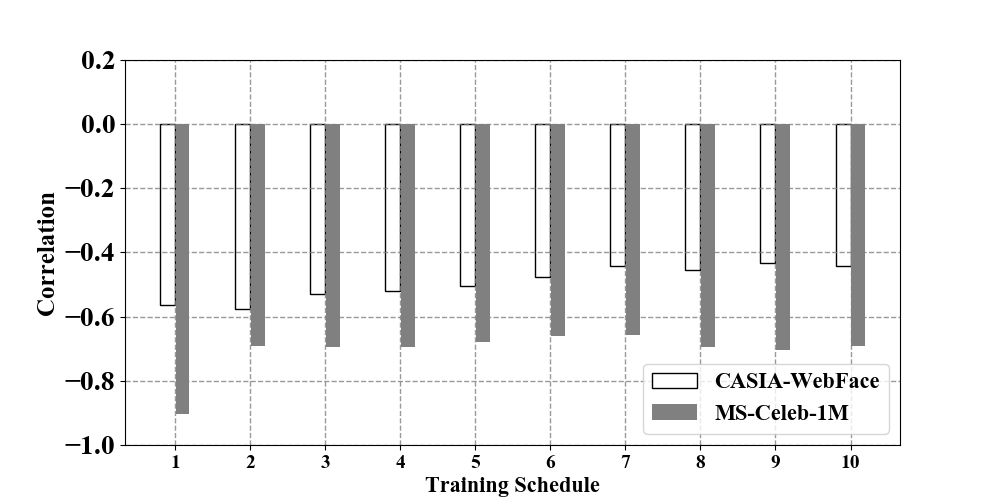}
\caption{Significant correlation between hard positive and hard negative throughout the training.
%on CASIA-WebFace and MS-Celeb-1M throughout the training.
}
\label{coorealtion}
\end{figure}

\subsection{Revisiting Margin-based Variants}
\label{subsec_revisit_variants}
Many prior works attempt to impose a margin in the positive logit to emphasize the supervision on the ground-truth class. Without loss of generality, we take the ArcFace formulation as an example. The positive logit $f_y = s \cos \left(\theta_{{i},{y}} + m\right)$ is equipped with a non-negative margin $m$, so the positive logit is decreased than the original version $f_y = s \cos \left(\theta_{{i},{y}}\right)$, as well as the probability: 

\begin{align}
    p_k | _{k = y} = 
    \frac{e^{s \cos \left(\theta_{{i},{y}} + m\right)}}
    {e^{s \cos \left(\theta_{{i},{y}} + m\right)} + \sum_{j=1,j\neq y}^{C} e^{s \cos \left(\theta_{{i},{j}}\right)}} \nonumber \\
    <
    \frac{e^{s \cos \left(\theta_{{i},{y}}\right)}}
    {e^{s \cos \left(\theta_{{i},{y}}\right)} + \sum_{j=1,j\neq y}^{C} e^{s \cos \left(\theta_{{i},{j}}\right)}}.
\label{proba_pos}
\end{align}

Then, according to Eqn.~\ref{gradient}, the supervision on the ground-truth class is amplified.
While the positive margin brings the benefit on ground-truth supervision, it will impair the non-ground-truth supervision. The reason is as follows. According to the property above-mentioned, the supervision on the non-ground-truth classes is emphasized by:
 
\begin{align}
    p_k | _{k \neq y} = 
    \frac{e^{s \cos \left(\theta_{{i},{k}}\right)}}
    {e^{s \cos \left(\theta_{{i},{y}} + m\right)} + \sum_{j=1,j\neq y}^{C} e^{s \cos \left(\theta_{{i},{j}}\right)}} \nonumber \\
    >
    \frac{e^{s \cos \left(\theta_{{i},{k}}\right)}}
    {e^{s \cos \left(\theta_{{i},{y}}\right)} + \sum_{j=1,j\neq y}^{C} e^{s \cos \left(\theta_{{i},{j}}\right)}}.
\label{proba_neg}
\end{align}
Unfortunately, such emphasis is activated to all the non-ground-truth classes indiscriminately. Thus, the hard non-ground-truth class, which deserved stronger supervision than the other classes, is, however, relatively weakened by the indiscriminate emphasis. More recently, MV-softmax, ArcNegFace and CurricularFace argue to perform an extra margin in the negative logits, and such scheme compensates the supervision on hard non-ground-truth class and alleviates the above issue. We take MV-softmax as an example, the logit of hard non-ground-truth class is reformulated as:
 
\begin{equation}
    f_j = s ( t \cos \left(\theta_{{i},{j}}\right) + t - 1   ),
\label{neg_logit_MV}
\end{equation}
where $t > 1$ can be regarded as the extra margin in the negative logit. We can see the logit is enlarged, and so the corresponding supervision is emphasised independently. 

\subsection{Improved Hard Negative Emphasis}
\label{Improved Hard Negative Emphasis}
In the above formulation (Eqn.~\ref{neg_logit_MV}) of non-ground-truth logit, the margin is implemented via the parameter $t$ in the negative logit. By further developing the gradient from Eqn.~\ref{gradient} with respect to the class weight $w_j$, 

\begin{align}
    \frac{\partial \mathcal{L}}{\partial w_j} | _{j \neq y}
    & =
    \frac{\partial \mathcal{L}}{\partial f_j}
    \frac{\partial f_j}{\partial w_j}
    \nonumber \\
    & =
    p_j \frac{\partial f_j}{\partial w_j}
    \nonumber \\
    & =
    p_j \frac{\partial s(t \cos \left(\theta_{{i},{j}}\right) + t - 1)}{\partial w_j}
    \nonumber \\
    & =
    p_j \frac{\partial s(t w_j^T x_i + t - 1)}{\partial w_j}
    \nonumber \\
    & =
    p_j s t x_i ,
\label{neg_grad_w}
\end{align}
we can see the supervision on the non-ground-truth class is determined by the predicted probability $p_k$ and the parameter $t$, while $p_k$ is also determined by $t$ (Eqn.~\ref{neg_logit_MV}). So, a slight increase of $t$ will lead to large increase of gradient, and thus bad solution or even unstable convergence (Fig.~\ref{flexible parameter}). But if we decrease $t$ to gain better convergence, the emphasis on the hard non-ground-truth class will be weakened instead. In order to alleviate the conflict between the stability and hard emphasis, we propose to disentangle the multiplicative margin and additive margin. To this end, the logit of non-ground-truth class is reformulated as:
 
\begin{align}
f_j^{M_{ij}}
= 
(1 - M_{{i},{j}}) \cdot s (\cos\theta_{{i},{j}}) 
% \nonumber \\
+ M_{{i},{j}} \cdot s (t\cos\theta_{{i},{j}}+\alpha),
\label{modulation_function}
\end{align}
where the mask $M_{{i},{j}} \in \{1, 0\}$ indicates whether the sample $x_i$ is hard to the ${j}$-th class. The choice of hard sample can be any of the off-the-shelf definition, such as mis-classification~\cite{wang2019mis}, OSM~\cite{wang2019deep}, DE-DSP~\cite{duan2019deep} \textit{etc.}  
More importantly, we disentangle the multiplicative margin and additive margin, and define them by $t$ and $\alpha$, respectively. 
$t$ and $\alpha$ represent the scale and the shift modulation. For hard samples, we emphasize the supervision on the hard non-ground-truth class by tuning $\alpha$ and $t$ together, so we can obtain the stable convergence while keeping hard supervision.
This is an improved formulation with more flexible parameter setting. One can refer to Section~\ref{subsec_exp_stable} and find our formulation leads to stable convergence.
%The scale and shift modulation should satisfy $t \geq 1$ and $\alpha \geq t - 1$ in order to guarantee $(t\cos(\theta_{{i},{j}})+\alpha)\geq \cos(\theta_{{i},{j}})$.

\subsection{Collaboration in Hard Positive}
\label{subsec_coop}
As discussed in Section~\ref{sec_intro}, when we train face recognition model on large-scale dataset, we can observe high correlation between hard positive and hard negative.
In this section, to further improve the training supervision, we explore to take advantage of the correlation between the hard positive and hard negative.
We argue that the margin formulated in the positive logit should be related to the negative logits for each sample. The hard samples are generally far away from their ground-truth class, and closer to the non-ground-truth classes. In other words, a sample which acts as hard case in positive perspective, also generally acts as hard one in negative perspective. We will discuss more about their correlation in Section \ref{neg and pos correlation}.
\par
Therefore, we develop an explicit collaboration between positive and negative logits for each sample. 
Specifically, a collaborative margin $\widetilde{m}_{i}$ is defined for the positive logit of the ${i}$-th sample.
Two factors are involved in the definition of collaborative margin: (1) the similarity to the non-ground-truth class $\cos{(\theta_{i,j})}|_{j \neq y}$ is involved to implement the collaboration; (2) the mask $M_{{i},{j}}$ is involved to enable the collaboration if it is a hard sample:
 
\begin{equation}
\begin{small}
\widetilde{m}_{i}=
\begin{cases}
 m_{0}\!+\! \frac {\sum_{j=1,j\neq y}^{C} (M_{{i},{j}}\cos{(\theta_{i,j})})}
 {\sum_{j=1,j\neq y}^{C}{M_{{i},{j}}}}{m}_{1}, & if ~~{\sum_{j=1,j\neq y}^{C}\!M_{{i},{j}}\!\neq\! 0}
 \\ \\
 m_{0}, & if ~~{\sum_{j=1,j \neq y}^{C}\!M_{{i},{j}}\! =\! 0}
\end{cases},
\label{mi}
\end{small}
\end{equation}
where $m_{0} > 0$ is a constant which maintains a basic margin for each sample, and $m_{1} > 0$ controls the range of the collaborative margin. 
We can see that the collaborative margin is related to the averaged hard negative logits. If the sample acts as a hard case from negative perspective, the collaborative margin $\widetilde{m}_{i}$ will increase; if it is not a hard case, $\widetilde{m}_{i}$ will reduce to the basic margin.

The collaborative margin can be applied in any positive-margin-based methods (\textit{e.g.},~\cite{liu2017sphereface,wang2018cosface,wang2018additive,deng2019arcface}). Here, we take ArcFace as an example, and the positive logit can be formulated as:
 
\begin{equation}
f_y^{M_{ij}}
= s \cos \left(\theta_{{i},{y}}+\widetilde{m}_{i}\right).
\end{equation}
When the sample has more hard negatives, then the collaborative margin $\widetilde{m}_{i}$ will increase, the positive logit will decrease and thus the loss value will increase.
Notice that each sample has its own collaborative margin $\widetilde{m}_{i}$ with respect to its hardness. 
In Section~\ref{subsec_hard_choice}, we will provide more discussion about the role of $M_{{i},{j}}$ in the collaborative margin.

Although the collaborative margin is recalculated in each iteration, it does not affect the computational cost significantly even when the training class number is large. This issue will be further discussed in Section~\ref{sec_training_efficiency}.
%it only increases a little computational cost even the training class number is large, which will be further discussed in Section~\ref{sec_training_efficiency}.
\par

\subsection{Negative-Positive Collaboration }
\label{subsec_npc_loss}
The Negative-Positive Collaboration (NPCFace) loss function incorporates the improved hard negative emphasis and the collaboration in hard positive emphasis, which is formulated as: 
 
\begin{equation}
\begin{small}
\!\mathcal{L}\!=\!-\frac{1}{N}\! \sum_{i=1}^{N}\! \log\! \frac{e^{f_y^{M_{ij}}}}{e^{f_y^{M_{ij}}}\! +\! \sum_{j=1,j \neq y}^{C} e^{ f_j^{M_{ij}}}}\!.
\end{small}
\end{equation}
As mentioned above, the choice of hard sample $M_{ij}$ can be any off-the-shelf definition, and we follow MV-softmax to employ the mis-classified samples. 
The collaboration comes from the important observation: a sample which is observed as a hard case in positive perspective, generally acts as hard one as well in negative perspective. 
So, NPCFace not only combines the emphasis from two views, but also benefits from the correlation between hard positive and hard negative for boosting the supervision.
The following sections will give more discussion on NPCFace and show its superiority on face recognition.

%------------------------------------------------------------------------
\section{Analysis}
\label{sec_analysis}
\subsection{Correlation Between Hard Positive and Negative}
\label{neg and pos correlation}

As mentioned above, the important observation is that a sample which is observed as a hard case from positive view, most likely will act as a hard case from negative view as well. This is the motivation of NPCFace that make use of this correlation for better training supervision.
To verify this argument, we calculate the correlation between the hard positives and hard negatives.
Specifically, we calculate the distance from the mis-classified samples to their ground-truth class, and the distance to the nearest non-ground-truth class; 
we calculate the correlation of the two distances of the samples, each of which has such two distance values;
we find the two distances are negatively correlated throughout the training (Fig.~\ref{coorealtion}). 
Note that this correlation is not the same item of the correlation between positive gradient and negative gradient in Section~\ref{subsec_revisit}.
Here, the correlation indicates the samples which have smaller distance to the non-ground-truth class (\textit{i.e.},  hard negative), will have larger distance to the ground-truth class (\textit{i.e.},  hard positive). Also, we can observe the correlation is more significant when the dataset has larger scale (MS-Celeb-1M is larger than CASIA-WebFace), which verifies the phenomenon in Fig.~\ref{hard_sample_scheme} (b).

\subsection{Discussion on Sample-wise Margin}
\label{subsec_samplewise}
The most prior works of margin-based methods, such as CosFace, AM-softmax, ArcFace \textit{etc.}, setup the margin with fixed value for all the training samples. Afterward, AdaptiveFace proposes to learn a margin for each class of the softmax classification. More recently, MV-softmax, ArcNegFace and CurricularFace set the margin in a sample-wise way, which means each training sample computes the loss with a specific margin with respect to the sample itself. This is a more reasonable routine because: (1) each training sample has different extent of hardness; (2) the hardness of a sample varies as the network being updated. Therefore, the sample-wise definition of margin is a better way. NPCFace also designs the margin in such sample-wise way, and employs this sample-wise routine for both positive margin and negative margin; however, MV-softmax, ArcNegFace and CurricularFace adopt the sample-wise margin only in the negative logits.

\subsection{Comparison with prior Pos-Neg routine}
\label{subsec_compare_arcneg_mv}
There are two prior counterparts to manipulate the hard positive and negative simultaneously in the training loss, \textit{i.e.},  MV-softmax and ArcNegFace. Whereas, NPCFace has different formulation and consequent advantages over them. 
(1) Firstly, MV-softmax, which does not have collaboration mechanism in it, deals with hard positive and hard negative independently.
(2) Secondly, ArcNegFace suppresses the supervision if the sample passes the decision boundary, only keeping focus on the boundary area and dropping the remaining samples out of the scope. By contrast, NPCFace always supplies adequate supervision from data as long as hard sample not handled yet by the learning network.
Besides, ArcNegFace has to apply together with label smoothing; otherwise, its performance will drop evidently~\cite{liu2019towards}. Label smoothing, however, could limit the model performance~\cite{zhang2020delving}. By contrast, NPCFace accomplishes the training without need of label smoothing.

%(2) Secondly, in ArcNegFace formulation, the supervision will be suppressed by the reduced negative logits, if the sample passes the decision boundary; that is to say, ArcNegFace only keeps focus on the boundary area and drops the remaining samples out of the scope. With such fixed focus, ArcNegFace cannot make full use of the label information during training with hard samples, whereas NPCFace always supplies adequate supervision from data as long as hard sample not handled yet by the learning network.
%(3) Thirdly, ArcNegFace has to apply together with label smoothing; otherwise, its performance will drop evidently~\cite{liu2019towards}. Label smoothing, however, could limit the model performance~\cite{zhang2020delving}, since the similarities between the given sample and its non-ground-truth categories are different. By contrast, NPCFace accomplishes the training without need of label smoothing.

%when the training sample is a hard case in positive perspective, then the scale modulation in the negative logits becomes small. That is to say, ArcNegFace argues to suppress the negative supervision (by reducing the negative logits) rather than emphasizing the negative supervision. Such practice is opposite of NPCFace, and is also contradictory to the high correlation between the hard positive and hard negative.
%Apparently, NPCFace designs a more reasonable routine to exploit the observed significant correlation (Fig.~\ref{coorealtion}) between hard positive and hard negative for training. 

\subsection{Analysis of Hard Sample}
\label{subsec_hard_choice}
\begin{figure}[t]
     %\centering
     %\hspace{-10pt}
     \begin{subfigure}[t]{0.24\textwidth}
         \centering
         \includegraphics[height=4.2cm]{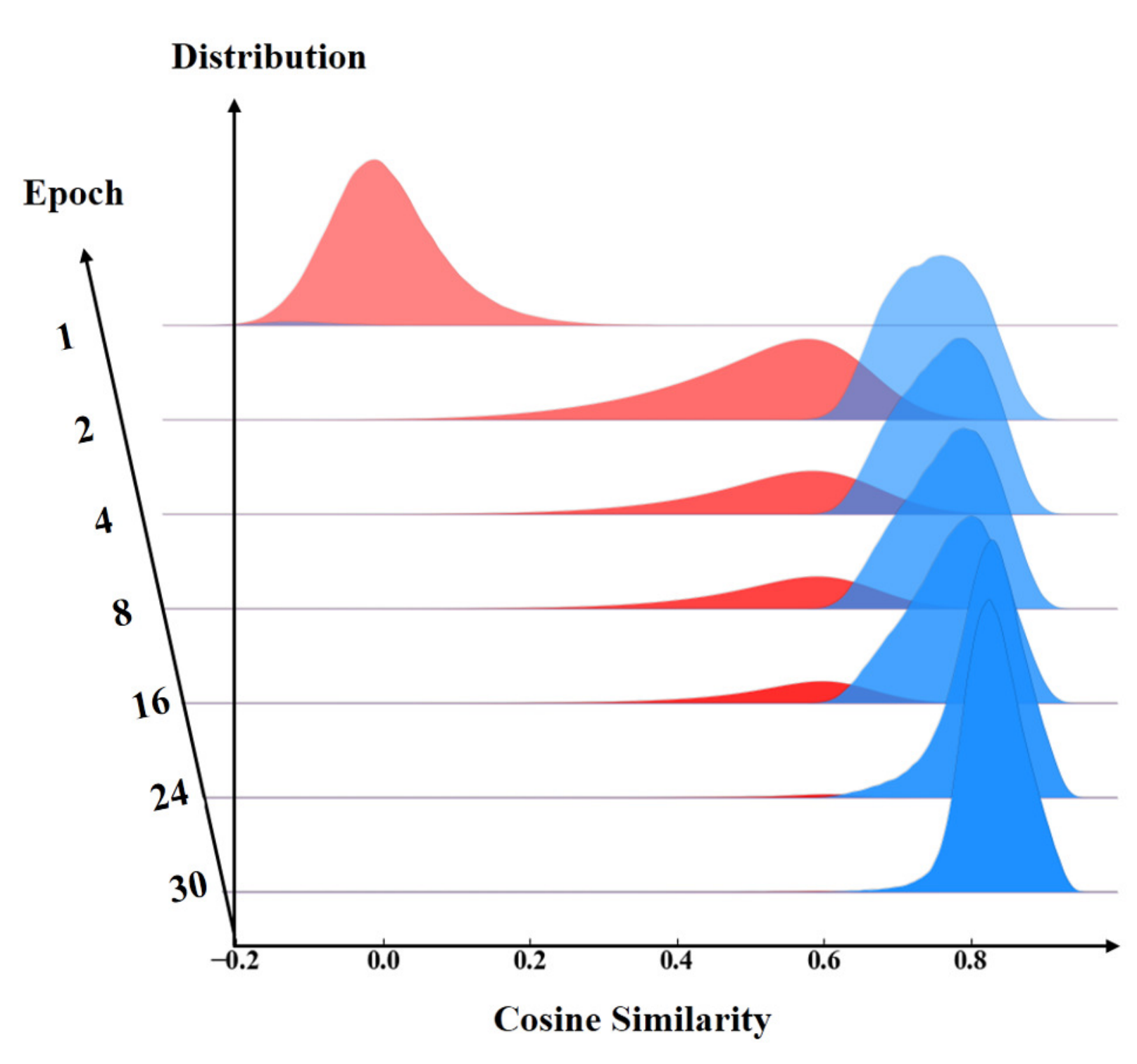}
         \caption{}
     \end{subfigure}
     \hfill
     %\hspace{-10pt}
     \begin{subfigure}[t]{0.24\textwidth}
         \centering
         \includegraphics[height=4.2cm]{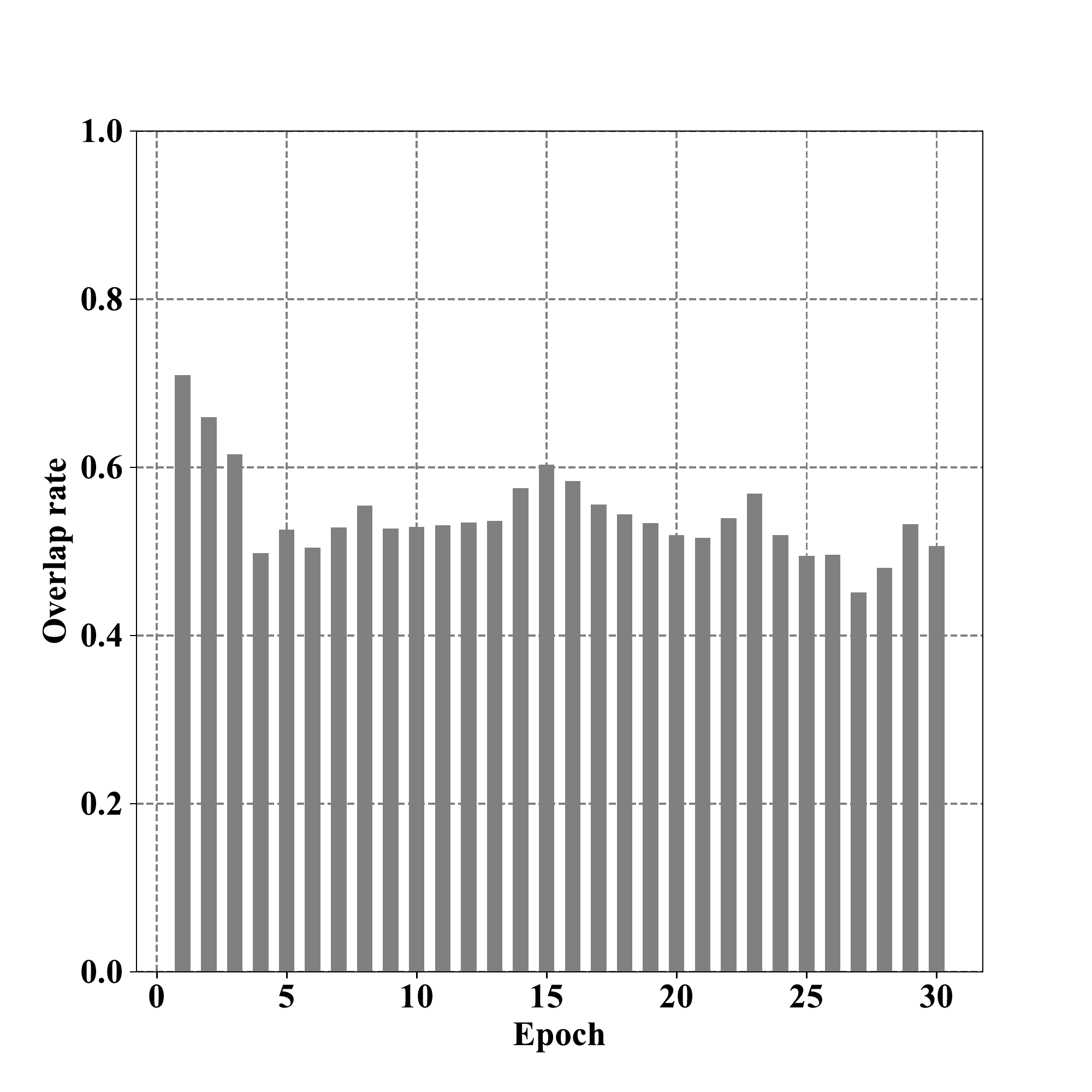}
         \caption{}
     \end{subfigure}

    \caption{
    (a) Blue: distribution of cosine similarity between well-classified samples and their ground-truth center throughout the training. Red: the counterpart between mis-classified samples and their ground-truth. Best viewed in color.
    (b) The overlap of the two distributions along with the training epochs.}
        \label{fig_distribution_overlap}
\end{figure}

There are many existing criteria for selecting hard samples for deep training.
%Support vector machine (SVM)~\cite{drucker1997support} aims at maximizing the margin between the support vectors of two classes. Considering support vectors are generally located around the decision boundary, it is reasonable to concentrate the training on the mis-classified samples which cross the decision boundary. 
In this paper, we choose the mis-classified samples as hard case rather than the one with large distance to the ground-truth center.
The mis-classified samples usually has smaller distance to the ground-truth than the well-classified one.
%Fig.~\ref{fig_svm} demonstrates the case in which the mis-classified sample has smaller distance to the ground-truth than the well-classified one.
As discussed in ~\cite{shi2016embedding}, this is caused by the highly-curved manifold in the feature space.
To verify this, we analyze the cosine similarity distribution between the training samples and their ground-truth centers throughout the training process.
In Fig.~\ref{fig_distribution_overlap} (a), the red distribution corresponds to the mis-classified samples, while the blue one corresponds to the well-classified samples. 
Their overlap rates are shown in Fig.~\ref{fig_distribution_overlap} (b).
At the start of training, almost all the samples are mis-classified because the network is trained from scratch.
Meanwhile, the overlap is the highest because the feature manifold is most distorted at this stage.
As the network gradually converging, the training samples become closer to their ground-truth centers. The red portion decreases and the blue portion increases because of less and less are mis-classified. Besides, we can observe that, as the network gradually converging, there is still an overlap between mis-classified samples and well-classified samples, which means it is improper if we directly use sample distance to identify hard samples. 
%This is the reason we take into account the mis-classification mask $M_{{i},{j}}$ in the collaborative margin (Eq.~\ref{mi}).
\par

\begin{comment}
\begin{figure}[t]
\centering
\includegraphics[scale=0.4]{./figures/svm.png}
\caption{An illustration of mis-classified case in the feature space. The distance from Class 2 center to mis-classified sample ($d_1$) is smaller than to a well-classified one ($d_2$).}
\label{fig_svm}
\end{figure}
\par
\end{comment}

\begin{comment}
\begin{figure}[t]
\centering
\includegraphics[scale=0.4]{./figures/experiment2-eps-converted-to.pdf}
\caption{Blue: distribution of cosine similarity between well-classified samples and their ground-truth center throughout the training. Red: the counterpart between mis-classified samples and their ground-truth. Best viewed in color.}
\label{fig_distribution}
\end{figure}
\par

\begin{figure}[t]
\centering
\includegraphics[width=8cm, height=6cm]{./figures/distance_iou-eps-converted-to.pdf}
\caption{The overlap of the two distributions along with the training epochs.}
\label{fig_overlap}
\end{figure}
\par
\end{comment}

\subsection{Robustness to Feature Dimension}
\label{subsec_dimension}
\begin{figure}[t]
\centering
\includegraphics[scale=0.4]{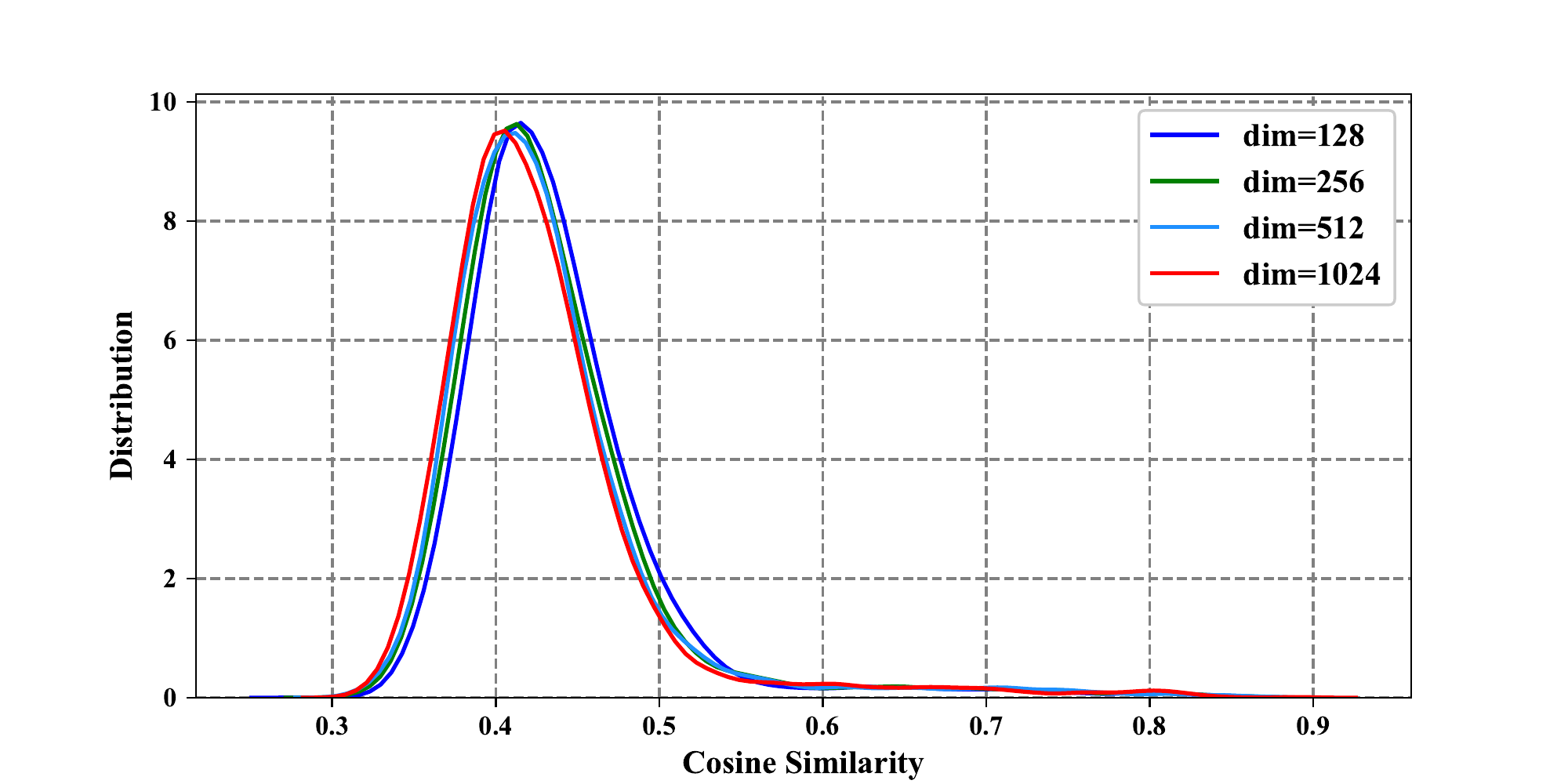}
\caption{Cosine similarity distribution between mis-classified samples and their nearest non-ground-truth centers.
The four networks output features of different dimensionality, but result in similar distributions.}
\label{distribution_neg}
\end{figure}

When embedding face images to feature space, the feature dimension plays an important role in metric computation. 
Here, we explore the effect of different dimension settings in NPCFace scheme. The output dimension of the last layer is set to 128, 256, 512 and 1,024 in four networks (with the same backbone), respectively. 
Then, the networks are trained with NPCFace. 
The cosine similarity distributions between mis-classified samples and their nearest negative centers are displayed in Fig.~\ref{distribution_neg}. We can observe that the hard negative similarity distributions are almost unchanged when the dimensionality increases from 128 to 1,024. 
The stability can be attributed to the margin formulation in the negative logit of NPCFace, which performs effective scaling and shifting in the training process.

\par

\section{Experiment}
\label{sec_exp}

This section is structured as follows. Section~\ref{subsec_exp_data} introduces the datasets and experimental settings. Section~\ref{subsec_exp_stable} studies the convergence of NPCFace and its flexibility on parameter setting. Section~\ref{subsec_exp_ablation} includes the ablation study which validates the negative margin, the collaborative positive margin and the combination. Section~\ref{subsec_exp_compare} demonstrates the comprehensive evaluation on a wide range of datasets and comparison with the state-of-the-art methods.

\subsection{Datasets and Experimental Setting}
\label{subsec_exp_data}

\par
\textbf{Training Data.}
We use two public datasets to train the networks. Specifically, we use cleaned CASIA-WebFace~\cite{yi2014learning} for training in stability analysis and ablation study, and we also utilize MS1M-v1c~\cite{trillionpairs.org} (cleaned version of MS-Celeb-1M~\cite{guo2016ms} ) for large scale comparison experiments. Note that we follow the lists of the~\cite{wang2019co} and~\cite{wang2019mis} to remove the overlapped identities between the employed training datasets and the test datasets. As a result, the CASIA-WebFace remains 9,879 identifies with 0.38M images and the MS1M-v1c remains 72,778 Identities and 3.28M images. 
\par
\textbf{Test Data.}
For a thorough evaluation, we use twelve test benchmarks, including LFW~\cite{huang2008labeled}, BLUFR~\cite{liao2014benchmark}, SSLFW~\cite{deng2017fine}, AgeDB-30~\cite{moschoglou2017agedb}, CFP-FP~\cite{sengupta2016frontal}, CALFW~\cite{zheng2017cross}, CPLFW~\cite{zheng2018cross}, RFW~\cite{wang2018racial}, MegaFace~\cite{kemelmacher2016megaface}, Trillion-Pairs~\cite{trillionpairs.org}, IJB-B~\cite{Whitelam2017IARPAJB}, IJB-C~\cite{Maze2018IARPAJB}. Among these test data, AgeDB-30 and CALFW focus on the large age gap face verification. CFP-FP and CPLFW aim at the large pose face verification. RFW focuses on the face verification for different races. SSLFW selects similar-looking face pairs to replace the negative pairs in LFW. 
BLUFR fully exploits all the LFW face images for the large-scale face recognition evaluation with focus at low FARs. MegaFace and Trillion-Pairs evaluate the performance of face recognition at the million scale of distractors. IJB-B and IJB-C contain images and videos for set-based face recognition. 
%------------------------------------------------------------------------
\par
\textbf{Preprocessing.} All face images are detected by FaceBoxes~\cite{zhang2017faceboxes}. 
Then, we align the faces by five facial landmarks~\cite{feng2018wing} and crop them to 120$\times$120 RGB.
During the training, we horizontally flip all the faces with probability 0.5 for data augmentation. Besides, each pixel in RGB images is normalized by subtracting 127.5 and then divided by 128.
\par
\begin{figure}[t]
\centering
%\hspace{-14mm}
\includegraphics[scale=0.45]{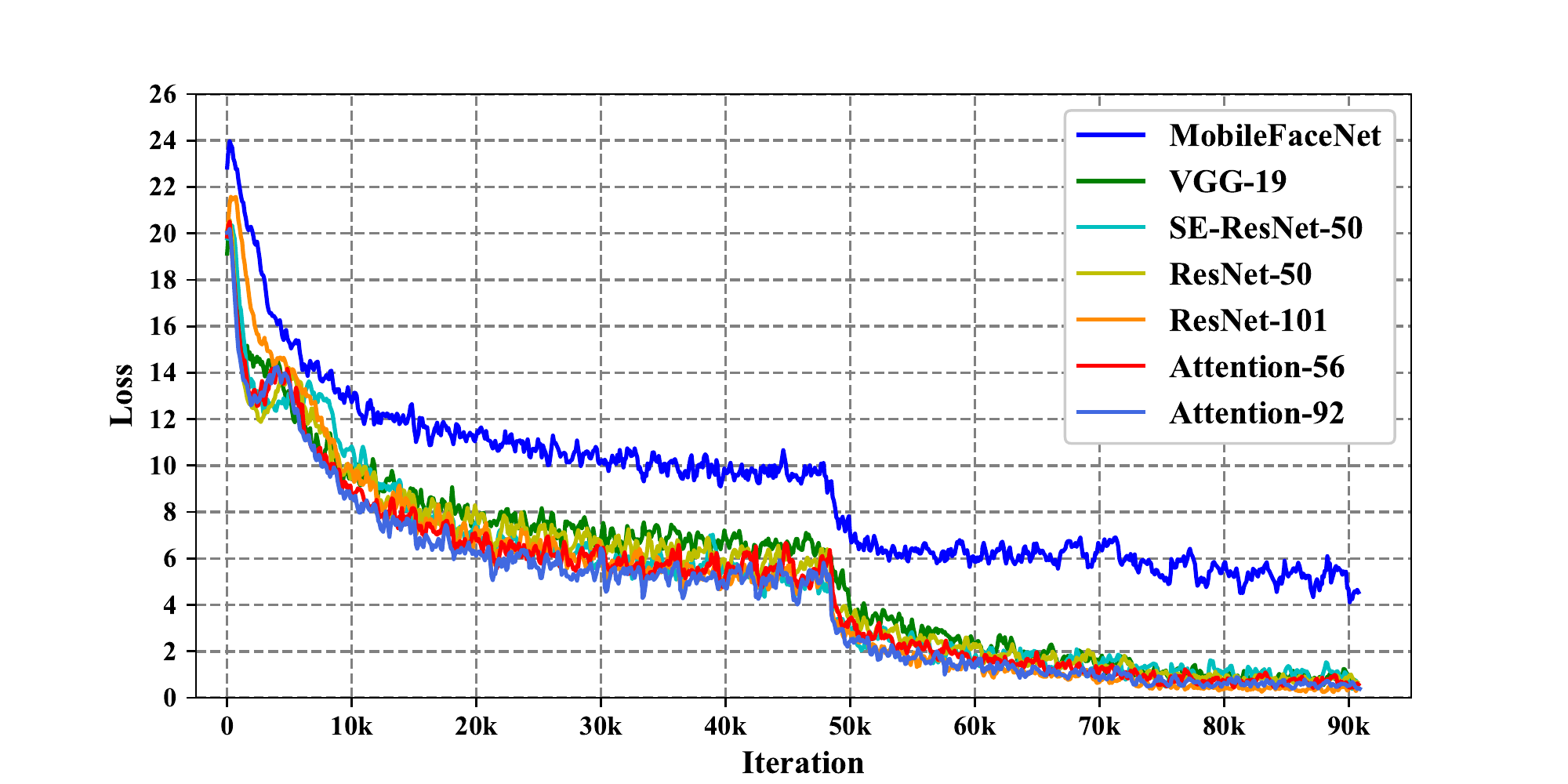}
\caption{ The loss value of NPCFace with different CNN architectures along training iterations. Best viewed in
color. }
\label{loss curves}
\end{figure}
\par

%------------------------------------------------------------------------
\textbf{CNN Architecture.} 
In the stability analysis and ablation study, we use  MobileFaceNet~\cite{chen2018mobilefacenets} as backbone to verify the effectiveness of each component of our method. Then, we adopt Attention-56~\cite{wang2017residual} as the backbone of NPCFace and all of the counterparts in the comparison experiments, so we can make a fair comparison while keeping the performance contrast between methods. 
%Compared to ResNet~\cite{he2016deep}, the Attention-56 can capture different types of attention to guide feature learning. 
%One can find more results with more heavy backbones in the supplementary material.
The output of network gives a 512-dimension feature. In addition, we also employ extra CNN architectures (Fig.~\ref{loss curves}), including VGG-19~\cite{Simonyan2014Very}, SE-ResNet-50~\cite{hu2018squeeze}, ResNet-50 and -101~\cite{he2016deep}, Attention-92~\cite{wang2017residual} to prove the convergence of our approach with various architectures. 
\par
%------------------------------------------------------------------------
\textbf{Training and Evaluation.} We train the networks from scratch on four NVIDIA Tesla P40 GPUs. On CASIA-WebFace, the batch size is 128 and the learning rate begins with 0.1 and is divided by 10 at the 16, 24, 28 epochs and finished at 30 epochs. On MS1M-v1c, we set the batch size as 512, and the learning rate starts form 0.1 and is divided by 10 at the 8, 14, 18 epochs and finish at 20 epochs. We set momentum to 0.9 and weight decay to 0.0005. According to the validation on LFW, we set $t = 1.1$ and $\alpha = 0.25$ in negative emphasis, and $m_{0}=0.4$ and $m_{1}=0.2$ in collaborative margin.
In the evaluation stage, we extract features from the last layer, and compute the cosine similarity as the similarity metric. 
For a fair and precise evaluation, all the overlapping identities between training and test datasets are removed according to the overlapping list~\cite{wang2019co} and~\cite{wang2019mis}. 
\par
\textbf{Compared Methods.} The original softmax is employed as baseline. 
The classification loss counterparts include SphereFace~\cite{liu2017sphereface}, CosFace~\cite{wang2018cosface}, ArcFace~\cite{deng2019arcface}, AdaM-softmax~\cite{liu2019adaptiveface}, AdaCos~\cite{zhang2019adacos}. In addition, we also compare with some recent softmax-based loss with hard mining improvement, such as MV-softmax~\cite{wang2019mis}, ArcNegFace~\cite{liu2019towards} and CurricularFace~\cite{Huang_2020_CVPR}. 
OHEM (HM-softmax~\cite{shrivastava2016training}) and Focal loss (F-softmax~\cite{lin2017focal}) are involved as the hard mining counterparts. Moreover, three other methods, including Triplet~\cite{schroff2015facenet}, Center loss~\cite{wen2016discriminative}, and UniformFace~\cite{duan2019uniformface}, are also employed for the comparison.
We re-implement them following every details in their original literature, and conduct fair comparison under the same experimental setting.

\begin{figure}[t]
     %\centering
     %\hspace{-10pt}
     \begin{subfigure}[t]{0.24\textwidth}
         \centering
         \includegraphics[height=3.5cm]{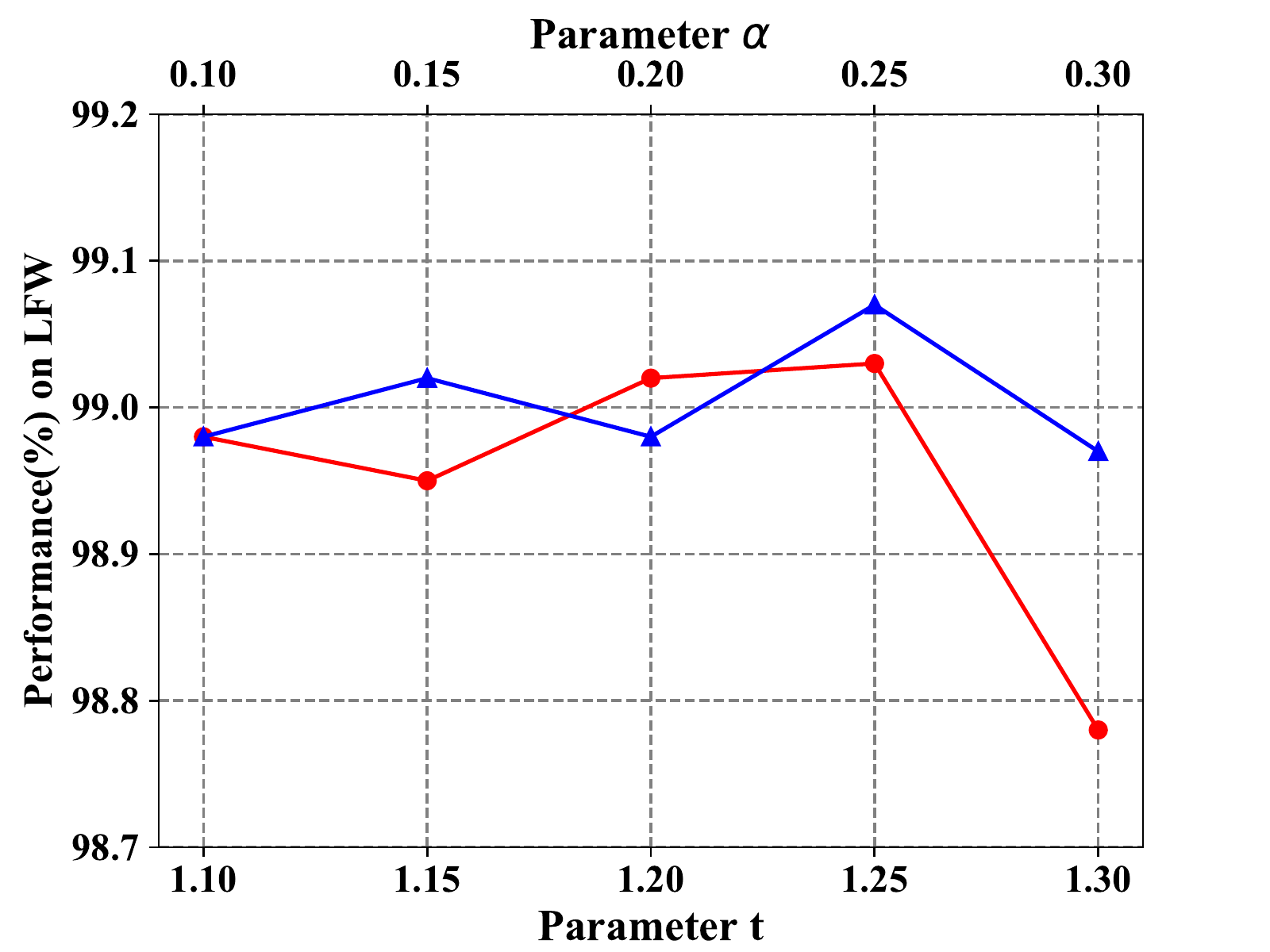}
         \caption{LFW}
     \end{subfigure}
     \hfill
     %\hspace{-10pt}
     \begin{subfigure}[t]{0.24\textwidth}
         \centering
         \includegraphics[height=3.5cm]{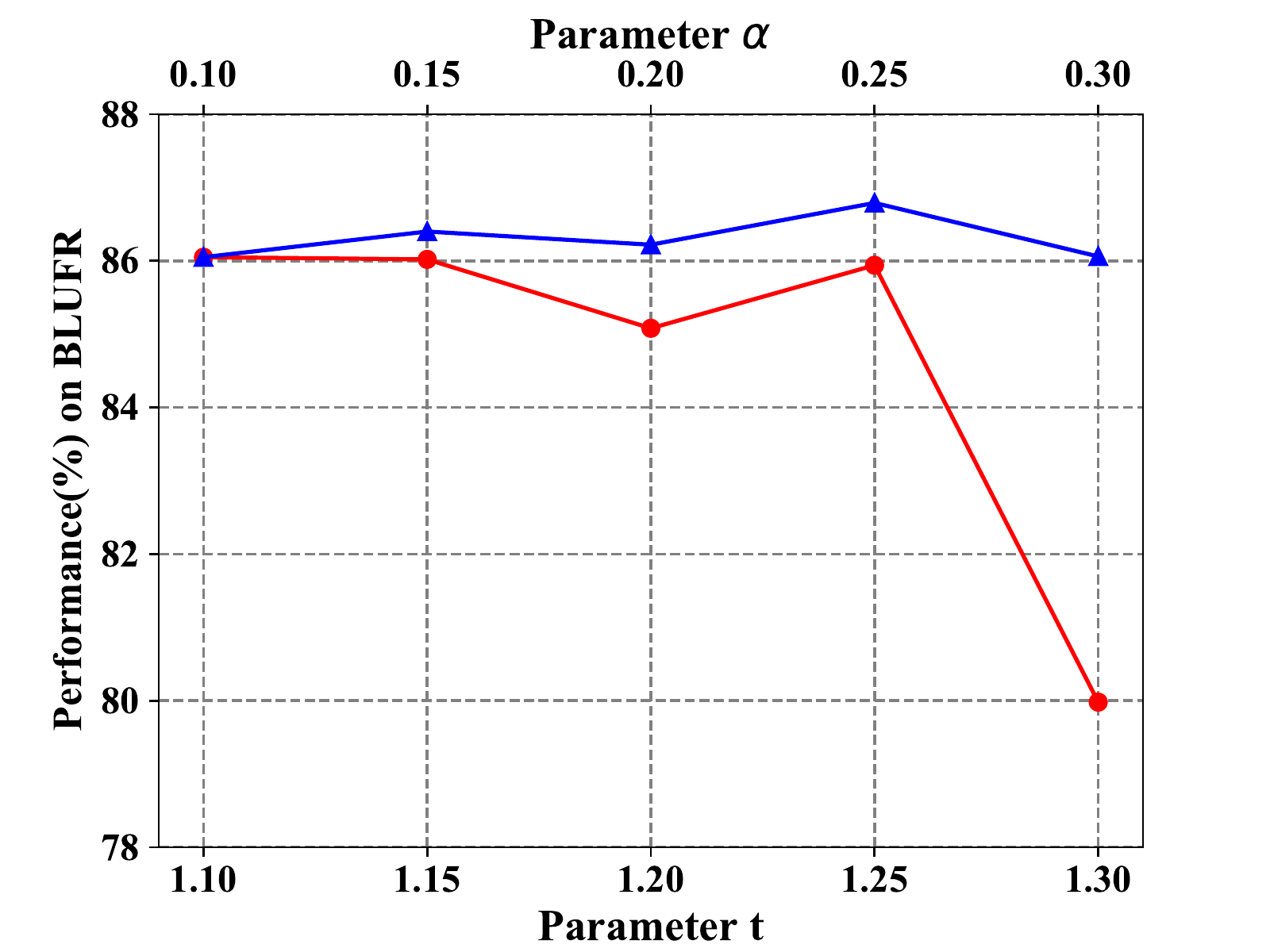}
         \caption{BLUFR}
     \end{subfigure}

    \caption{
    The accuracy on LFW and BLUFR with different $\it{t}$ and $\alpha$. 
    The blue denotes NPCFace, and the red denotes the counterpart MV-softmax.
    Best viewed in color.}
        \label{flexible parameter}
\end{figure}

\begin{comment}
\begin{figure}[t]
\centering
\hspace{-20mm}
\begin{minipage}[t]{0.6\linewidth}
\includegraphics[height=5cm]{./figures/Performance_of_negative_lfw-eps-converted-to.pdf}
\subcaption{LFW}
\label{flexible_lfw}
\end{minipage}
\par
\vspace{1em}
\hspace{-20mm}
\begin{minipage}[t]{0.6\linewidth}
\includegraphics[height=5cm]{./figures/Performance_of_negative_blufr-eps-converted-to.pdf}
\subcaption{BLUFR}
\label{flexible_blufr}
\end{minipage}
\caption{The results on LFW and BLUFR with different $\it{t}$ and $\alpha$. The red line is the performance of MV-softmax with different $\it{t}$. The blue line is the performance of NPCFace with different $\alpha$ and $t=1.1$. Best viewed in color.}
\label{flexible parameter}
\end{figure}
\end{comment}

\subsection{Stable Convergence and Flexible Setting}
\label{subsec_exp_stable}

To demonstrate the stable convergence in the training, we employ NPCFace to train on five prevailing CNN architectures, including MobileFaceNet~\cite{chen2018mobilefacenets}, VGG-19~\cite{Simonyan2014Very}, ResNet-50 and -101~\cite{he2016deep}, SE-ResNet-50~\cite{hu2018squeeze} and Attention-56 and -92~\cite{wang2017residual}. As illustrated in Fig.~\ref{loss curves}, the loss values gradually drop along with training iterations. To demonstrate the flexible parameter setting of our improved formulation in the negative logits, we conduct a comparison experiment with NPCFace and MV-softmax. As shown in Fig.~\ref{flexible parameter}, the red line is the performance of MV-softmax and blue line is the NPCFace. We can find there is a large decrease both in LFW and BLUFR (VR@FAR=1e-5) when $t=1.3$ for MV-softmax, because MV-softmax's shift parameter is entangled with $t$. But for NPCFace, we can fix $t$ in an appropriate range (\textit{e.g.},~ $t = 1.1$) and enlarge $\alpha$ to obtain further performance improvement. So, NPCFace is more flexible to determine favorable training parameters.

\begin{table}[t]
\setlength{\tabcolsep}{1pt}
\begin{center}
\caption{Ablation study: performance ($\%$) on BLUFR and MegaFace. On BLUFR, we report the verification rate at FAR of 1e-4 and 1e-5. On MegaFace, ``Id.'' refers to face identification rank-1 accuracy with 1M distractors, and ``Veri.'' refers to face verification TAR at 1e-6 FAR.}
\label{ablation}
\begin{tabular}{|c|c||p{1.25cm}<{\centering}|p{1.25cm}<{\centering}|p{1.25cm}<{\centering}|p{1.25cm}<{\centering}|}
\hline
\multirow{2}{*}{~Neg.~}&
\multirow{2}{*}{~Pos.~}&
\multicolumn{2}{c|}{BLUFR}&
\multicolumn{2}{c|}{MegaFace}
\\\cline{3-6}&&1e-4&1e-5 & Id. &Veri.\\
\hline\hline
- & -&92.74&83.52&72.89&77.64\\
\checkmark & -&94.31&86.84&77.49&80.86\\
- &\checkmark &94.29&86.31&75.33&79.58\\
\hline
\checkmark&\checkmark &\textbf{94.82}&\textbf{88.15}&\textbf{77.76} &\textbf{82.29}\\
\hline
\end{tabular}
\end{center}
\end{table}

\subsection{Ablation Study}
\label{subsec_exp_ablation}

In this subsection, we analyse the two improvements of NPCFace and validate their effectiveness. 
Table~\ref{ablation} shows the results on BLUFR and MegaFace. The baseline (\textit{i.e.},  the top row) is the original ArcFace~\cite{deng2019arcface}. ``Neg." represents the employment of our improved negative logits. The improvement by negatives (second row) is significant by every evaluation metric. 
``Pos." refers to the collaborative margin for positive logit. We also observe the obvious improvement (third row) compared with the baseline. By the joint advantage of the two components, NPCFace (bottom row) can obtain further performance improvement, especially at the low FARs.

%------------------------------------------------------------------------
\begin{table*}[t]
\begin{center}
\caption{Performance ($\%$) comparison on the LFW, SLLFW, AgeDB-30, CFP-FP, CALFW, CPLFW and RFW.}
\label{LFW}
\centering
\resizebox{\textwidth}{!}
{
\begin{tabular}{|c|p{1cm}<{\centering}|p{1cm}<{\centering}|p{1cm}<{\centering}|p{1cm}<{\centering}|p{1.25cm}<{\centering}|p{1.25cm}<{\centering}|c|c|c|c|c|}
\hline
\multirow{2}{*}{Method}&
\multirow{2}{*}{LFW}&
\multirow{2}{*}{SLLFW}&
\multirow{2}{*}{AgeDB}&
\multirow{2}{*}{CFP-FP}&
\multirow{2}{*}{CALFW}&
\multirow{2}{*}{CPLFW}&
\multicolumn{4}{c|}{RFW}
\\ \cline{8-11}&&&&&&&Caucasian&Indian &Asian &African\\
\hline\hline
softmax 
&99.45&98.53&96.58&92.67&93.52&86.27&95.35&91.63&87.80&89.45\\
\hline
HM-softmax~\cite{shrivastava2016training}
&99.67&98.87&96.43&93.33&94.02&86.95&94.77&90.65&87.35&87.47\\
F-softmax~\cite{lin2017focal}
&99.65&98.73&96.60&94.11&93.87&87.17&94.95&90.72&86.82&88.00\\
\hline
Center loss~\cite{wen2016discriminative} 
&99.65&98.82&96.83&93.37&94.23&86.58&96.03&92.07&89.60&90.85\\
Triplet loss~\cite{schroff2015facenet}
&99.58&98.42&96.27&92.30&93.27&85.07&95.65&87.03&82.55&85.88\\
UniformFace~\cite{duan2019uniformface}
&99.70&99.18&96.90&94.34&94.40&87.45&96.90&93.08&90.37&92.00\\
\hline
SphereFace~\cite{liu2017sphereface}
&99.70&98.97&96.43&93.86&94.17&87.81&95.95&91.95&89.72&90.48\\
CosFace~\cite{wang2018cosface}
&99.73&99.33&97.53&94.83&95.07&88.63&97.98&94.93&93.80&94.88\\
ArcFace~\cite{deng2019arcface} 
&99.75&99.38&97.68&94.27&95.12&88.53&98.22&95.68&93.97&94.95\\
AdaCos~\cite{zhang2019adacos}
&99.68&99.00&97.15&94.03&94.38&87.03&97.37&92.00&90.15&91.92\\
AdaM-softmax ~\cite{liu2019adaptiveface}
&99.74&99.37&97.68&94.96&95.05&88.80&98.22&95.13&93.77&94.58\\
MV-AM-softmax~\cite{wang2019mis} &99.72&99.40&97.73&93.77&95.23&88.65&98.28&95.08&93.50&94.57\\
ArcNegFace~\cite{liu2019towards}
&99.73&99.28&97.37&93.64&95.15&87.87&98.07&95.73&93.35&95.05\\
CurricularFace~\cite{Huang_2020_CVPR}
&99.72&99.33&97.43&93.73&94.98&87.62&98.23&95.37&93.60&94.73\\
\hline
NPCFace 
&\textbf{99.77}&\textbf{99.57}&\textbf{97.77}&\textbf{95.09}&\textbf{95.60}&\textbf{89.42}&\textbf{98.58}&\textbf{95.98}&\textbf{94.78}&\textbf{95.52}
\\
\hline
\end{tabular}}
\end{center}
\end{table*}

\subsection{Comparison Experiments}
\label{subsec_exp_compare}

The comparison experiments aims to evaluate NPCFace against various challenges, and show the results compared with state-of-the-art methods.
\par
\textbf{Recognition against large pose and age gap.}
Table~\ref{LFW} includes the performance on LFW, CPLFW, CFP-FP, CALFW and AgeDB-30. For LFW evaluation, NPCFace has a small improvement, since state of the art on LFW is almost saturated. 
The other four benchmarks (CPLFW, CFP-FP, CALFW and AgeDB-30) aim at the evaluation when encountering hard cases of large face pose and large age gap.
From the results, we can observe that NPCFace is better than the baseline softmax loss and other competitors in all evaluation, which prove the effectiveness of our negative-positive collaboration.
\par
%------------------------------------------------------------------------
\textbf{Recognition against similar looking.}
By the leading accuracy on SLLFW (second column in Table~\ref{LFW}) which contains many similar-looking identities, NPCFace also shows its advantage of dealing with hard cases of inter-class.

\par
%------------------------------------------------------------------------

\textbf{Recognition for various races.} 
The RFW benchmark includes four testing subsets, \textit{i.e.}, Caucasian, Asian, Indian and African. Each contains about 3,000 individuals with 6,000 image pairs for face verification. 
As shown on right half of Table~\ref{LFW}, NPCFace achieves the highest accuracy in the four testing subsets, especially in the challenging subsets of Asian and African. It indicates the good generalization ability of NPCFace training for various races.
%------------------------------------------------------------------------
\par
%\textbf{Results on BLUFR, MegaFace and Trillion-Pairs.} 
\textbf{Recognition at low FAR.} 
Table~\ref{MF_nd_TP} includes the performance at low FARs.
First, we conduct the evaluation on BLUFR protocol and compare the verification rate at FAR of 1e-4 and 1e-5.
We can observe that our method is obviously superior to all the competitors. Further, we compare the performance in the MegaFace Challenge, which is one of the most challenging benchmark for large scale face identification and verification. Following the official protocol, we use FaceScrub~\cite{ng2014data} as the probe set. 
Compared with the baseline softmax loss, our method achieves at least 4 percents improvements on both the Rank-1 identification rate and the verification TAR@FAR=1e-6. Compared with the recent state-of-the-art methods (CosFace, ArcFace, MV-AM-softmax, AdaCos, AdaM-softmax and ArcNegFace \textit{etc.}), our method also keeps the superiority, which proves the effectiveness of the collaborative margins. The Trillion-Pairs Challenge~\cite{trillionpairs.org} is also a large scale face recognition challenge, which is consisted of 5,700 identities for recognition and 1.58 million faces as distractors. Table~\ref{MF_nd_TP} also displays the performance comparison in the Trillion-Pairs Challenge. From the results, we find that the hard mining methods~\cite{shrivastava2016training,lin2017focal} do not work well in the extreme low FAR range (\textit{i.e.}, 1e-9), while the margin-improved methods (NPCFace, ArcFace, MV-AM-softmax \textit{etc.}) shows the advantage on exploiting hard samples. Besides, we observe that our NPCFace is able to push the limit of deep face recognition in the extreme low FAR range and achieve the leading performance among all the competitors both in identification and in verification.

\begin{figure}[t]
     %\centering
     %\hspace{-10pt}
     \begin{subfigure}[t]{0.24\textwidth}
         \centering
         \includegraphics[height=4cm]{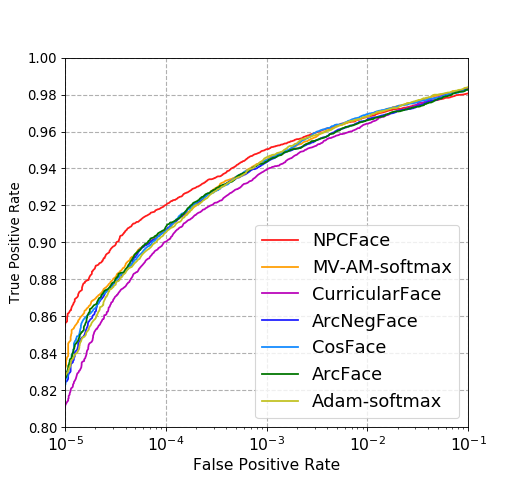}
         \caption{IJB-B}
     \end{subfigure}
     \hfill
     %\hspace{-10pt}
     \begin{subfigure}[t]{0.24\textwidth}
         \centering
         \includegraphics[height=4cm]{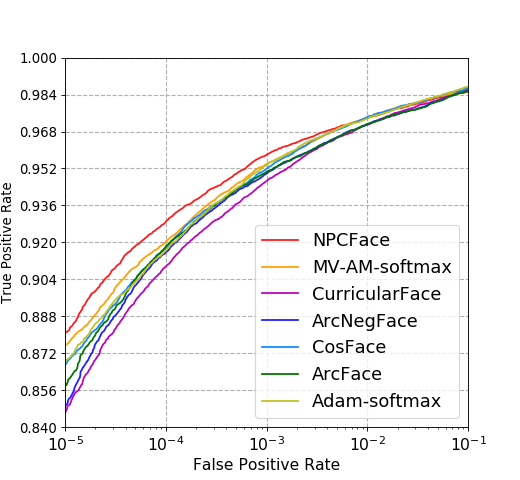}
         \caption{IJB-C}
     \end{subfigure}

    \caption{
    The ROC curves of NPCFace and the counterparts on IJB-B and IJB-C. Best viewed in color.
    }
        \label{IJB_ROC}
\end{figure}

In addition, we also report the performance of face verification task on IJB-B~\cite{Whitelam2017IARPAJB} and IJB-C~\cite{Maze2018IARPAJB} datasets. The IJB-B dataset consists of 1,845 subjects with 21,798 images and 55K video frames. For face verification task, there are 10,270 positive matches and 8M negative matches. The IJB-C dataset composes of 3,531 subjects with 31,334 images and 117,542 video frames, which provides 19,557 genuine matches and 15,638,932 impostor matches. For the IJB-B and IJB-C face verification evaluation, we obtain the set-based representations by averaging the image features without any specific strategies for set-based face recognition.
From the results in Table.~\ref{IJB} and Fig.~\ref{IJB_ROC}, we can find NPCFace also keep the leading performance on IJB-B and IJB-C datasets, which shows our methods can obtain more discriminate and generalized features than the counterparts.

\begin{table}[t]
\setlength{\tabcolsep}{4.8pt}
\begin{center}
\caption{Performance ($\%$) comparison on BLUFR, MegaFace and Trillion-Pairs.}
\label{MF_nd_TP}
% \resizebox{\linewidth}{!}
{\begin{tabular}{|c|c|c|c|c|c|c|}
\hline
\multirow{2}{*}{Method}&
\multicolumn{2}{c|}{BLUFR}&
\multicolumn{2}{c|}{MegaFace}&
\multicolumn{2}{c|}{Trillion-Pairs}
\\ \cline{2-7}&1e-4&1e-5 & Id.&Veri.&Id.&Veri.\\
\hline\hline
softmax &99.43&97.62&92.27&93.77&51.21&47.91\\
\hline
HM-softmax~\cite{shrivastava2016training} 
&99.50&97.77&91.45&93.51&49.78&46.66\\
F-softmax~\cite{lin2017focal} 
&99.58&97.40&91.59&92.93&45.69&41.58\\
\hline
{Center loss}~\cite{wen2016discriminative} &99.59&98.12&93.91&94.80&-&-\\ 
{Triplet loss}~\cite{schroff2015facenet}&95.83&77.56&86.70&91.61&-&-\\
UniformFace~\cite{duan2019uniformface}  & 99.60&98.16&94.13&95.21&-&- \\
\hline
SphereFace~\cite{liu2017sphereface}
&99.51&98.03&92.54&94.23&55.09&54.42\\
CosFace ~\cite{wang2018cosface}
&99.79&98.73&96.65&97.25&72.33&70.98\\
ArcFace~\cite{deng2019arcface}
&99.80&98.53&97.04&97.38&75.68&74.80\\
AdaCos~\cite{zhang2019adacos}
&99.63&97.44&94.27&96.04&53.59&52.33\\
AdaM-softmax~\cite{liu2019adaptiveface}
&99.81&98.89&96.80&97.37&71.76&70.70\\
MV-AM-softmax~\cite{wang2019mis}
&99.81&99.25&97.13&97.50&75.34&74.34\\
ArcNegFace~\cite{liu2019towards}
&99.78&98.49&96.85&97.35&75.48&73.77\\
CurricularFace~\cite{Huang_2020_CVPR}
&99.79&98.85&96.80&97.24&75.07&73.45\\
\hline
NPCFace
&\textbf{99.83}&\textbf{99.36}&\textbf{97.75}&\textbf{98.07}&\textbf{77.53}&\textbf{77.01}\\
\hline
\end{tabular}}
\end{center}
\end{table}

\begin{table}[t]
\begin{center}
\caption{Performance ($\%$) comparison on IJB-B and IJB-C.}
\label{IJB}{
\begin{tabular}{|c|c|c|c|c|}
\hline
\multirow{2}{*}{Method}&
\multicolumn{2}{c|}{IJB-B}&
\multicolumn{2}{c|}{IJB-C}
\\ \cline{2-5}&1e-4&1e-5&1e-4&1e-5\\
\hline\hline
softmax &85.66&73.63&86.62&76.48\\
\hline
HM-softmax~\cite{shrivastava2016training} 
&85.81&73.79&87.26&77.76\\
F-softmax~\cite{lin2017focal} 
&85.10&73.67&86.98&77.53\\
\hline
Center loss~\cite{wen2016discriminative} &86.43&74.16&86.87&76.64\\
Triplet loss~\cite{schroff2015facenet} &73.21&40.37&78.12&48.07\\
UniformFace~\cite{duan2019uniformface}  & 87.22&75.01& 88.87&79.64\\
\hline
SphereFace~\cite{liu2017sphereface}
&86.67&74.75&87.92&78.77\\
CosFace ~\cite{wang2018cosface}
&90.60&82.28&91.72&86.68\\
ArcFace~\cite{deng2019arcface}
&90.83&82.68&91.82&85.75\\
AdaCos~\cite{zhang2019adacos}
&86.04&73.34&87.53&78.91\\
AdaM-softmax~\cite{liu2019adaptiveface}
&90.54&82.70&91.64&86.84\\
MV-AM-softmax~\cite{wang2019mis}
&90.67&83.17&92.03&87.52\\
ArcNegFace~\cite{liu2019towards}
&90.62&81.59&90.91&85.64\\
CurricularFace~\cite{Huang_2020_CVPR}
&90.04&81.15&90.95&84.63\\
\hline
NPCFace
&\textbf{92.02}&\textbf{85.59}&\textbf{92.90}&\textbf{88.08}\\
\hline
\end{tabular}}
\end{center}
\end{table}

\begin{comment}
\begin{figure}[t]
\centering
% \hspace{-20mm}
\begin{minipage}[t]{0.42\linewidth}
\includegraphics[height=3.5cm]{./figures/roc_ijbb.png}
\subcaption{IJB-B}
\label{IJB-B}
\end{minipage}
\hfill
\begin{minipage}[t]{0.42\linewidth}
\includegraphics[height=3.5cm]{./figures/roc_ijbc.png}
\subcaption{IJB-C}
\label{IJB-C}
\end{minipage}
\caption{The ROC curves of NPCFace and the counterparts on IJB-B and IJB-C. Best viewed in color.}
\end{figure}
\end{comment}

\subsection{Training Efficiency}
\label{sec_training_efficiency}
During the training process with NPCFace, a sample-wise margin is calculated in each iteration, which increases a little more computational cost with a large number of training classes. To verify it, we make a comparison experiment on the training efficiency of the loss functions. Specifically, we utilize CASIA-WebFace and MS-Celeb-1M as the training datasets, respectively. The Attention-56 is employed as the backbone. As shown in Table~\ref{training_efficiency}, all the margin-based loss functions require a bit more time cost than the original softmax. Moreover, we can observe the extra time cost of NPCFace is negligible on CASIA-WebFace. For large number of training classes (\textit{i.e.}, MS-Celeb-1M), the time cost of NPCFace increases only 6.1$\%$ compared with CosFace and ArcFace. The result can demonstrate our method only increases a small piece of computational cost in the training with large number of classes.

\begin{table}[t]
    %\arrayrulecolor{blue}
    \begin{center}
    %\color{blue}
    \caption{ Comparison of NPCFace and the existing loss functions in terms of training efficiency. ``sec./batch'' refers to the average time for one forward-backward propagation per mini-batch (the batch size is set as 512). 
    The experiment is performed on four NVidia P40 GPUs for each loss function.}
    \label{training_efficiency}
    \vspace{0.2cm}
    \begin{tabular}{|c|c|c|}
    \hline
         Method & \makecell[c]{CASIA-WebFace\\ $\sim$ 10k IDs\\(sec./batch)}&  \makecell[c]{MS-Celeb-1M\\ $\sim$ 100k IDs\\(sec./batch)} \\
    \hline\hline
    softmax &1.03&1.14 \\
    \hline
    HM-softmax~\cite{shrivastava2016training} &1.03&1.14 \\
    F-softmax~\cite{lin2017focal} &1.03&1.14\\
    \hline
    Center loss~\cite{wen2016discriminative}  &1.05&1.21\\
    Triplet loss~\cite{schroff2015facenet}  &1.28&1.28\\
    UniformFace~\cite{duan2019uniformface}  &1.05&1.22\\
    \hline
    SphereFace~\cite{liu2017sphereface}
    &1.05&1.20\\
    CosFace ~\cite{wang2018cosface}
    &1.04&1.15\\
    ArcFace~\cite{deng2019arcface}
    &1.04&1.15\\
    AdaCos~\cite{zhang2019adacos}
    &1.04&1.21\\
    AdaM-softmax~\cite{liu2019adaptiveface}
    &1.04&1.15\\
    MV-AM-softmax~\cite{wang2019mis}
    &1.04&1.15\\
    ArcNegFace~\cite{liu2019towards}
    &1.04&1.21\\
    CurricularFace~\cite{Huang_2020_CVPR}
    &1.04&1.19\\
    \hline
    NPCFace&1.05&1.22\\
    \hline
    \end{tabular}
    \end{center}
    \end{table}

\section{Conclusion}
\label{sec_concl}
In this paper, we propose a novel training supervision, namely Negative-Positive Collaboration (NPCFace) loss, to address the challenges in large-scale face recognition. The contribution consists in two folds.
First, a collaborative training emphasis on hard positives and hard negatives is developed to make full use of them for better training.
Second, the improved margin formulation in the negative logits leads to stable convergence and flexible parameter setting.
The two components can jointly bring advantages to the training of deep face recognition. Consequently, NPCFace achieves favorable performance in the low FAR range and various hard situations, and shows it superiority over the prior methods. 

\section*{Acknowledgement}
This work is supported by the National Key R\&D Program of China under Grant No.2020AAA0103800.

\ifCLASSOPTIONcaptionsoff
  \newpage
\fi



%\bibliographystyle{IEEEtran}

\bibliographystyle{IEEEtran}
\bibliography{egbib}



% \end{thebibliography}


%iography text here.
%\end{IEEEbiography}


%\newpage



%\vfill


\end{document}